\pdfoutput=1

\documentclass[11pt]{article}

\usepackage{emnlp2021}

\usepackage{times}
\usepackage{latexsym}

\usepackage[T1]{fontenc}

\usepackage[utf8]{inputenc}

\usepackage{microtype}

\usepackage{url}            
\usepackage{booktabs}       
\usepackage{amsmath}
\usepackage{amsfonts}       
\usepackage{nicefrac}       
\usepackage{natbib}
\usepackage{array}
\usepackage{makecell}
\usepackage{multirow}
\usepackage{amsfonts}
\usepackage{graphicx}
\usepackage{wrapfig} 
\usepackage{float}
\usepackage{caption} 
\usepackage{mathtools}
\usepackage[autostyle]{csquotes}
\usepackage{xcolor}
\usepackage{caption}
\usepackage{subcaption}
\usepackage{multirow}
\usepackage{enumitem}

\newcommand{\KL}[2]{\mathbb{D}_{\text{KL}}\left[ #1 || #2 \right]}
\DeclareMathOperator*{\E}{\mathbb{E}}
\newcolumntype{M}[1]{>{\centering\arraybackslash}m{#1}}
\newcolumntype{N}{@{}m{0pt}@{}}
\newcolumntype{P}[1]{>{\centering\arraybackslash}p{#1}}

\colorlet{myGreen}{green!40!gray}
\newcommand{\NoGap}{\vspace{-5mm}}
\newcommand{\tablegap}{\vspace{0.5em}}

\newcommand{\qfull}{q_{\phi}(\hat{r}_{1:K}|r_{1:N}, s)}

\newcommand{\prior}{p(\hat{r}_{1:K} | r_{1:N})}
\newcommand{\decoder}{p_{\theta}(s|\hat{r}_{1:K})}

\newcommand{\ROUGE}{\textsc{R1 top-k}}
\newcommand{\ours}{\textsc{SelSum}}
\newcommand{\dataset}{\textsc{AmaSum}}
\newcommand{\pc}{pros and cons}

\newcommand{\rsep}{\!$\vert \vert$ \!}

\makeatletter
\newcommand{\thickhline}{%
    \noalign {\ifnum 0=`}\fi \hrule height 1pt
    \futurelet \reserved@a \@xhline
}
\makeatother

\title{Learning Opinion Summarizers by Selecting Informative Reviews}

\author{Arthur Bražinskas$^1$, Mirella Lapata$^1$ \and Ivan Titov$^{1, 2}$ \\
$^1$ ILCC, University of Edinburgh  \\
$^2$ ILLC, University of Amsterdam \\
\texttt{abrazinskas@ed.ac.uk, \{mlap,ititov\}@inf.ed.ac.uk} \\
}

\date{}

\begin{document}

\setlength{\belowdisplayskip}{10pt}
\setlength{\abovedisplayskip}{10pt}

\maketitle

\begin{abstract}
    
Opinion summarization has been traditionally approached with unsupervised, weakly-supervised and few-shot learning techniques. In this work, we collect a large dataset of summaries paired with user reviews for over 31,000 products, enabling supervised training. However, the number of reviews per product is large (320 on average), making summarization -- and especially training a summarizer -- impractical. Moreover, the content of many reviews is not reflected in the human-written summaries, and, thus, the summarizer trained on random review subsets hallucinates. In order to deal with both of these challenges, we formulate the task as jointly learning to select informative subsets of reviews and summarizing the  opinions expressed in these subsets. The choice of the review subset is treated as a latent variable, predicted by a small and simple selector.
The subset is then fed into a more powerful summarizer. 
For joint training, we use amortized variational inference and policy gradient methods. Our experiments demonstrate the importance of selecting informative reviews resulting in improved quality of summaries and reduced hallucinations. 
\end{abstract}

\section{Introduction}
\label{sec:introduction}


\begin{table}[t!]
    \centering
 	\footnotesize 
    \begin{tabular}{ >{\centering\arraybackslash} m{1cm} m{5cm}} 
    \hline
    \textbf{Verdict} & \vspace{0.8em} 
     If you like the idea of a \textcolor{red}{glass feeder}, this is the one to get. It has \textcolor{orange}{a lot to offer for the price}.
    \vspace{0.8em} \\ \hline
    \textbf{Pros} & \vspace{0.8em}  
        $\bullet$ Has a \textcolor{blue}{large opening} that makes it \textcolor{blue}{easy to get in and out} of the feeder \hspace{0.25em} \vspace{0.3em}  \newline 
        $\bullet$ Has a \textcolor{cyan}{nice design} that's \textcolor{myGreen}{easy to clean}
        \vspace{0.8em}  \\ \hline
  \textbf{Cons} & \vspace{0.8em} 
  $\bullet$ The \textcolor{brown}{lid is a little flimsy}, and it's \textcolor{purple}{not as durable as some of the other models} 
  \vspace{0.8em}  \\ \hline
    \textbf{Reviews} & \vspace{0.8em} 
    ... looks just as nice as the \textcolor{red}{glass feeders} \rsep ... Very happy with the \textcolor{orange}{value, quality and function} ... \rsep ... \textcolor{orange}{the cheapest most flexible "jar"} I've ever seen ... \rsep ... \textcolor{blue}{Nice large opening} so it's easy to pour the sugar water \rsep ... This feeder has a nice \textcolor{blue}{large opening} ... \rsep ... this is the \textcolor{cyan}{perfect design} and size ... \rsep \textcolor{blue}{The hummingbirds liked it and had no trouble feeding or perching}.... \rsep ... The main compartment is \textcolor{myGreen}{easy to clean}... \rsep ... \textcolor{brown}{The top is a little flimsy} ... \rsep
    ... \textcolor{purple}{it fell out of the hanger it broke for good} ... \textcolor{purple}{there are so many other nice ones out there that have glass "jar's" or at least sturdier plastic} ... \rsep  ... \textcolor{myGreen}{The tray is easy to clean} ...
  \vspace{1.0em}  \\ \hline
    \end{tabular}
    \caption{Example summary generated by \ours{} with colored alignment to the input reviews. The reviews are truncated and delimited with `\rsep'.}
    \label{table:front-example-summ}
    \NoGap
\end{table}

Summarization of user opinions expressed in online resources, such as blogs, reviews, social media, or internet forums, has drawn much attention due to its potential for various information access applications, such as creating digests, search, and report generation~\citep{hu2004mining,medhat2014sentiment,angelidis2018summarizing, amplayo2019informative}. Although significant progress has been observed in supervised summarization in non-subjective context, such as news articles~\citep{rush2015neural, nallapati2016abstractive, see2017get, lebanoff2018adapting, gehrmann-etal-2018-bottom, fabbri2019multi, laban2020summary}, modern deep learning methods rely on large
amounts of annotated data that are not readily available in the
opinion-summarization domain and expensive to produce. Specifically, the annotated datasets range from 50 to 200 annotated products \citep{chu2019meansum, brazinskas2020-unsupervised, angelidis2018summarizing, angelidis2020extractive}.

The absence of large high-quality resources for supervised learning has called for creative solutions in the past. There is a long history of applying unsupervised and
weakly-supervised methods to opinion summarization \citep{mei2007topic, titov2008modeling,
  angelidis2018summarizing, chu2019meansum, amplayo2020unsupervised, brazinskas2020-unsupervised}.

\begin{figure*}
     \centering
     \includegraphics[width=0.75\textwidth]{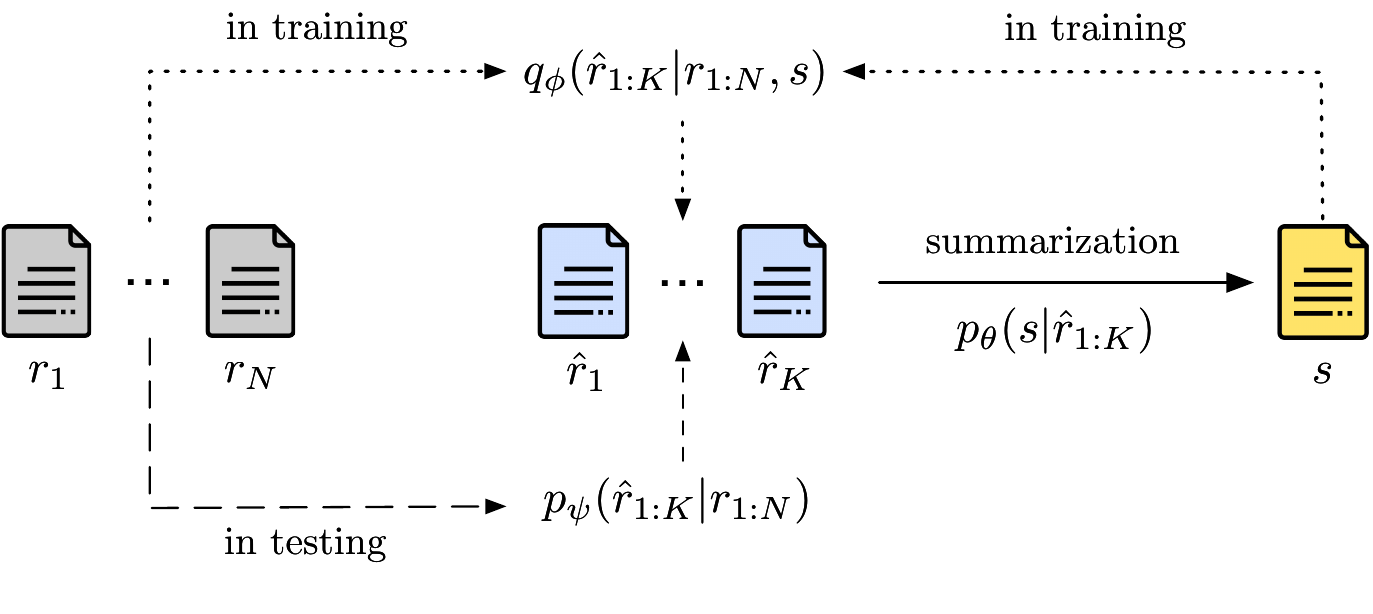}
     \vspace{-1em}
     \caption{The \ours{} model is trained to select and summarize a subset of relevant reviews $\hat r_{1:K}$ from a full set $r_{1:N}$ using the approximate posterior $q_{\phi}(\hat{r}_{1:K}|r_{1:N}, s)$. To yield review subsets in test time, we fit and use a parametrized prior $p_{\psi}(\hat r_{1:K}|r_{1:N})$.}
     \label{fig:model}
      \vspace{-3mm}
 \end{figure*}

In this work, we introduce the largest multi-document opinion summarization dataset \dataset{} consisting of verdicts, \pc{} for more than 31,000 summarized Amazon products, as shown in Table~\ref{table:front-example-summ}. The summaries were written by professional product reviewers guiding online audience to make better purchasing decisions. In turn, each product is linked to more than 320 reviews, on average. This, however, makes it virtually impossible to train a conventional encoder-decoder model using standard hardware.
Moreover, not all reviews cover the summary content. Thus, training to predict summaries based on random review subsets results in hallucinations, as we will empirically demonstrate in Sec.~\ref{sec:content_support}. This calls for specialized methods selecting smaller subsets of relevant reviews that are fed to a summarizer. We explore this direction by introducing \ours{} that jointly learns to \textbf{se}lect and \textbf{sum}marize review subsets using \textit{amortized variational inference} and \textit{policy gradient optimization} \citep{kingma2013auto, mnih2014neural, deng2018latent}, as depicted in Fig.~\ref{fig:model}.

To select relevant review subsets in training, we utilize the summary to pre-compute lexical features. Then we score review relevance with a tiny neural selector that has only 0.1\% of the deep encoder's parameters. These simple features, as opposed to deep encoder representations, allow us to select reviews from large collections without a significant computational burden. Subsequently, only selected reviews are encoded by an `expensive' encoder, in order to predict the summary. To select quality review subsets in test time, when the summary is not available, we approximate the summary relevance using another neural selector. In our experiments, we show the importance of accurate review selection, affecting the summarizer in training and its output in testing. Furthermore, we show that our model outperforms alternatives in terms of ROUGE scores and content fidelity. All in all, our contributions can be summarized as follows\footnote{The codebase and dataset are available at \url{https://github.com/abrazinskas/SelSum}.}:
\begin{itemize}
    \item We provide the largest dataset for multi-document opinion summarization;
    \item We propose an end-to-end model selecting and summarizing reviews; 
    \item We empirically demonstrate superiority of our model to alternatives.
\end{itemize}

\section{Dataset}
\label{sec:dataset}

The dataset (\dataset{}) is based on summaries for consumer products written by professional reviewers, in English. We focused on four main professional product review platforms: \texttt{bestreviews.com} (BR); \texttt{cnet.com}; \texttt{pmag.co.uk} (PM); \texttt{runrepeat.com} (RR). The former three mostly offer content for electronic consumer products, while the last one for sport shoes. These summaries provide a quick-glance overview of a product to help users make informed purchases. Unlike customer reviewers on public platforms, such as Amazon, professional reviewers concentrate on quality writing and deliberately utilize many sources of information. These sources include reading customer reviews on public platforms, making online research, asking expert users for an opinion, and testing products themselves.  
In general, the summaries come in two forms. The first ones are verdicts, usually a few sentences emphasizing the most important points about a product. The second ones are \pc{}, where the most important positive and negative details about a product are presented. These tend to be more detailed, and focus on  fine-grained product aspects, such as Bluetooth connectivity, resolution, and CPU clock speed. 


As content providers compete for online users, the summaries \textbf{are} what the user wants as opposed to what researchers \textbf{believe} the user wants. This is in contrast to crowd-sourcing where researchers bias the worker writing process with assumptions about what constitutes a good summary. The assumptions are rarely verified by a marketing research or user testing. In turn, this has lead to a large variance of summary styles and composition even in the same domain \citep{angelidis2018summarizing, chu2019meansum, brazinskas2020-unsupervised, brazinskas2020few}.

\subsection{Content Extraction}
We wrote HTML scraping programs for each platform and 
extracted segments containing verdicts, and \pc{}. Further, from advertisement links we extracted Amazon standard identification numbers (ASINs) which allowed us to identify what Amazon products are reviewed and link summaries to the Amazon product catalog. 

We used various paid services to obtain Amazon reviews and product metadata. We fetched verified reviews for all products, and utilized unverified ones only for unpopular products ($<$ 50 reviews). We also utilized a publicly available Amazon review dataset~\citep{ni2019justifying}.

\subsection{Filtering}
We removed all reviews that have less than 10 and more than 120 words. We also removed all unpopular products that have less than 10 reviews. Further, we removed all summaries that have less than 5 words, and all instances that have either verdict or pros or cons missing. The overall statistics comparing our final dataset to available alternatives are shown in Table ~\ref{table:comparative_statistics}. Our dataset is substantially larger than the alternatives, both in terms of number of summaries and their associated reviews.

\begin{table*}[h]
    \centering
    \begin{tabular}{ l | c | c | c | c | c}
        & Ent & Rev/Ent & Summaries (R) & Type & Domain \\
        \thickhline
        \dataset{} (This work) & 31,483 & 326 & 33,324 (1.06) & Abs. & Products\\
        \textsc{Space} \citep{angelidis2020extractive} & 50 & 100 & 1,050 (3) & Abs. & Hotels \\
        \textsc{Copycat} \citep{brazinskas2020-unsupervised} & 60 & 8 & 180 (3) & Abs. & Products \\
        \textsc{FewSum} \citep{brazinskas2020few} & 60 & 8 & 180 (3) & Abs. & Businesses \\
        \textsc{MeanSum} \citep{chu2019meansum} & 200 & 8 & 200 (1) & Abs. & Businesses \\
        \textsc{OpoSum} \citep{angelidis2018summarizing} & 60 & 10 & 180 (3) & Ext. & Products \\
        \thickhline
    \end{tabular}
    \caption{Statistics comparing our dataset to alternatives; \textsc{R} stands for the number of references. For our dataset, we show the average number of reviews and references per entity. We count verdicts, \pc{} of a product as one summary. }
    \label{table:comparative_statistics}
    \NoGap{}
\end{table*}

\subsection{Summary Statistics}
\label{sec:summary_statistics}

\begin{table*}[h!]
    \centering 
    \begin{tabular}{l | c c c | c c c | c c c }
    \multicolumn{1}{c}{} & \multicolumn{3}{c}{\textbf{Verdict}} & \multicolumn{3}{c}{\textbf{Pros}} & \multicolumn{3}{c}{\textbf{Cons}}\\\thickhline
     & Len & R1 & R2 & Len & R1 & R2 & Len & R1 & R2 \\
    \thickhline
    BR (27,329) & 20.60 & 82.40 & 34.45 & 37.34 & 79.12 & 29.75 & 16.27 & 82.19 & 33.58 \\
    CNET (2,717) & 29.74 & 81.05 & 34.72 & 32.08 & 77.85 & 30.04 & 25.11 & 75.16 & 25.84 \\
    PM (1,756) & 30.23 & 76.08 & 28.28 & 20.78 & 65.53 & 16.09 & 14.33 & 62.08 & 13.81 \\
    RR (1,522) & 77.86 & 60.45 & 13.12 & 120.04 & 59.44 & 13.47 & 43.36 & 63.11 & 16.02 \\
    \thickhline
    All (33,324) & 24.47 & 80.95 & 33.18 & 39.82 & 77.40 & 28.31 & 18.12 & 79.69 & 31.10 \\
    \thickhline
    \end{tabular}
    \caption{Summary statistics of the dataset. The number of data points is in parentheses.}
    \label{table:summary_stats_full}
\end{table*}

We analyzed summaries from different platforms in terms of their lengths and ROUGE recall with respect to reviews, as shown in Table~\ref{table:summary_stats_full}. First of all, verdicts tend to be shorter than \pc{}, and concentrate on fewer aspects. They also exhibit higher word overlap to user reviews as indicated by higher ROUGE scores. We also observed that \pc{} tend to concentrate on specific product features, which can often be found in product meta information (product description, the bullet list of features). Cons tend to be shorter than pros, we believe, primarily because most summarized products are rated highly (4.32/5.0 on average). 


\section{Approach}
\label{sec:approach}
As summaries are written mostly for popular products, with more than 320 reviews on average,
it is computationally challenging to encode and attend all the available ones to decode the summary. To alleviate this problem, we propose to condition the decoder on a smaller subset of reviews. However, as not all reviews provide a good content coverage of the summary. Thus, training on random subsets leads to hallucinations, as we will show in Sec.~\ref{sec:content_support}. Instead, we propose to learn a review selector, which chooses reviews guided by the summary. We frame this as a latent variable modeling problem (the selection is latent) and rely on the  variational inference framework to train the selector, see Sec.~\ref{sec:model}. The selector (the approximate posterior) is a neural module assessing the review-summary relevance using pre-computed lexical features, thus, efficiently selecting from large review collections. Further, the selected reviews are decoded to the summary, as illustrated in Fig.~\ref{fig:model}. To select reviews in test time, we train a review selector that does not rely on the summary, as presented in Sec.~\ref{sec:fitting_prior}.


\subsection{Probabilistic Framing}
\label{sec:probabilistic_framing}
Let $\{r_{1:N}^i, s^i\}_{i=1}^M$ be reviews-summary pairs, and let $\hat{r}_{1:K}$ be a reduced subset of reviews, where $K < N$, and each variable follows a categorical distribution. As review subsets $\hat{r}_{1:K}$ are unknown in advance, they are latent variables in our model, and both the full set $r_{1:N}$ and the summary $s$ are observed variables. To maximize the log-likelihood shown in Eq.~\ref{eq:log_likelihood}, we have to marginalize over all possible review subsets.
\begin{equation}
\begin{aligned}
    &\log p_{\theta}(s| r_{1:N}) = \log  \E_{\hat{r}_{1:K} \sim p(\hat{r}_{1:K} | r_{1:N})} \left[p_{\theta}(s| \hat{r}_{1:K}) \right]
    \label{eq:log_likelihood}
\end{aligned}
\end{equation}
Unfortunately, the marginalization is intractable, and thus we leverage the Jensen's inequality \citep{boyd2004convex} to obtain the lower bound as shown in Eq.~\ref{eq:lower_bound_prior}, which, in turn, is approximated via Monte Carlo (MC).
\begin{equation}
\begin{aligned}
  &\log \E_{\hat{r}_{1:K} \sim p(\hat{r}_{1:K} | r_{1:N})} \left[  \decoder \right] \geq \\ &\E_{ \hat{r}_{1:K} \sim \prior} \left[ \log \decoder \right]
  \label{eq:lower_bound_prior}
\end{aligned}
\end{equation}
Here the latent subset $\hat r_{1:K}$ is sampled from a prior categorical distribution agnostic of the summary. From the theoretical perspective, it can lead to a large gap between the log-likelihood and the lower bound, contributing to poor performance \citep{deng2018latent}. From the practical perspective, it can result in the input reviews not covering the summary content, thus forcing the decoder in training to predict `novel' content. Consequently, this leads to hallucinations~\citep{maynez2020faithfulness} in test time, as we empirically demonstrate in Sec.~\ref{sec:content_support}.

\vspace{-2mm}
\subsection{Model}
\label{sec:model}

To address the previously mentioned problems, we leverage \textit{amortized inference} reducing the gap~\citep{kingma2013auto, cremer2018inference}. And re-formulate the lower bound using weighted sampling as shown in Eq.~\ref{eq:amortized_lower_bound}.
\begin{equation}
\begin{aligned}
  &\log \E_{\hat{r}_{1:K} \sim p(\hat{r}_{1:K} | r_{1:N})} \left[  p_{\theta}(s|\hat{r}_{1:K}) \right] \geq \\ 
&\E_{\hat{r}_{1:K} \sim q_{\phi}(\hat{r}_{1:K}|r_{1:N}, s)} \left[ \log p_{\theta}(s|\hat{r}_{1:K})\right] - \\ &\KL{q_{\phi}(\hat{r}_{1:K}| r_{1:N}, s)}{p(\hat{r}_{1:K}| r_{1:N})} 
\label{eq:amortized_lower_bound}
\end{aligned}
\end{equation}
The first term, known as \textit{reconstruction}, quantifies the summary prediction quality with review subsets selected by the approximate posterior $\qfull$. Unlike the prior, it selects reviews relevant to the summary $s$, thus providing a better content coverage of the summary. Hence, it reduces the amount of `novel' content the decoder needs to predict. As we empirically demonstrate in Sec.~\ref{sec:content_support}, this results in summaries with substantially fewer hallucinations in test time.
The second term, the Kullback-Leibler divergence (KLD), serves as a regularizer preventing the posterior from deviating from the prior. We did not find it useful -- presumably because the latent space of our model
 (i.e. the choice of reviews to summarize) has already very limited capacity -- and do not use the KLD term in training. Instead, after training, we fit a rich prior
 (see Sec.~\ref{sec:fitting_prior}). 
\subsubsection{Approximate Posterior}
\label{sec:approximate_posterior}
 
The distribution assigns a probability to every possible subset of reviews $\hat r_{1:K}$. However, 
this would require us to consider $N!/(N-K)!K!$ possible combinations to normalize the distribution \citep{koller2009probabilistic}. To make it computationally feasible, we assume a local, left-to-right factorization~\citep{larochelle2011neural}, reducing the complexity to $\mathcal{O}(K\, N)$: 
\begin{equation}
    q_{\phi}(\hat{r}_{1:K}|r_{1:N}, s) = \prod_{k=1}^K q_{\phi}(\hat{r}_k|r_{1:N}, \hat{r}_{1:k-1}, s).
    \label{eq:approximate_posterior_factorization}
\end{equation}
Technically, each local distribution can be computed by softmax normalizing scores produced by the \textit{inference network} $f_{\phi}(\hat r_k, r_{1:N}, s)$. To represent ($\hat r_k$, $r_{1:N}$, $s$) input tuples, we use pre-computed lexical features, such as ROUGE scores for ($\hat r_k$, $s$) and ($\hat r_k$, $r_{1:N}$), and aspect-coverage metrics (see~Appendix~\ref{app:posterior_features} and Section ~\ref{sec:posterior_selected_review_subsets}). This, in turn, allows us to learn feature inter-dependencies and score large collections of reviews in a fast and memory efficient-manner.

To avoid duplicate reviews, we assume that $\hat{r}_k$ can be any review in the full collection $r_{1:N}$ except previously selected ones in the partial subset $\hat{r}_{1:k-1}$. To accommodate that, we `block' scores for all previously selected reviews $\hat r_{1:{k-1}}$ as $f_{\phi}(\hat r_k, r_{1:N}, s) = - \inf $ $\forall{\hat r_k \in \hat r_{1:{k-1}}}$. In practice, we compute logits once for $r_{1:N}$, and then perform a progressive distribution re-normalization by `blocking' logits for previously selected reviews. 

\subsubsection{Reconstruction}
\label{sec:rec_term}
In training, we optimize parameters only for the reconstruction term in Eq.~\ref{eq:amortized_lower_bound}. However, this optimization is not straightforward as it requires backpropagation through categorical samples $\hat r_{1:K}$ to compute a gradient estimate. Furthermore, it is not possible to apply the re-parametrization trick \citep{kingma2013auto} for categorical variables. On the other hand, the Gumbel-Softmax trick \citep{jang2016categorical}, in its standard form, would require encoding and backpropagating through all possible review subsets, making it computationally infeasible. Instead, we used REINFORCE \citep{williams1992simple} that  considers only a sampled subset for gradient estimation,\footnote{We provide further discussion contrasting REINFORCE and the Gumbel-Softmax trick in Appendix~\ref{app:RI_vs_gs}.} as shown in Eq.~\ref{eq:rec_grad}. The notation is simplified to avoid clutter.
\begin{equation}
    \begin{aligned}
        &\nabla_{\phi} \E_{\hat r_{1:K} \sim q_{\phi}(\hat{r}_k|r_{1:N}, \hat{r}_{1:k-1}, s)} \left[ \log p_{\theta}(s|\hat r_{1:K})\right] = \\ &\E_{\hat r_{1:K} \sim q_{\phi}} \left[ (\log p_{\theta}(s|\hat r_{1:K}) - \beta(s)) \nabla_{\phi} \log q_{\phi}\right]
    \end{aligned}
    \label{eq:rec_grad}
\end{equation}
Here $\beta(s)$ corresponds to a baseline reducing the gradient variance \citep{greensmith2004variance}. Specifically, we used an MC estimate of Eq.~\ref{eq:lower_bound_prior} by randomly sampling review subsets. Moreover, we were separately updating the posterior and summarizer, in the spirit of stochastic inference \citep{hoffman2013stochastic}. In turn, this made it computationally possible to further reduce the variance by estimating Eq.~\ref{eq:rec_grad} with more samples. 

\subsection{Fitting a Prior}
\label{sec:fitting_prior}

The selector used in training (i.e the approximate posterior $q_{\phi}(\hat{r}_{1:K}|r_{1:N}, s)$)  cannot be used in test time, as it has a look-ahead to the summary $s$.  Instead we need a prior $p_{\psi}(\hat{r}_{1:K}|r_{1:N})$. 
Since we have not used any prior in training (i.e. ignored the KLD term, Eq.~\ref{eq:amortized_lower_bound}), we, similarly in spirit to \citet{razavi2019generating}, fit a parameterized prior after training the summarizer, and then use the prior as the test-time review selector.

Intuitively, the fitted prior tries to mimic predictions of the approximate posterior without having access to $s$. We care only about the mode of the distribution, so, to simplify the task, we select the most likely review subset from the posterior to train the test time selector and frame it as a binary prediction task. Let $\{r_{1:N}^i, s^i\}_{i=1}^M$ be reviews-summary pairs where we utilize $\qfull$ to create $\{r_{1:N}^i, d_{1:N}^i\}_{i=1}^M$ pairs. Here, $d_j$ is a binary tag indicating whether the review $r_j$ was selected by the posterior. This dataset is then used to train the score function $f_{\psi}(r_k; r_{1:N})$. In test time, we select $K$ reviews with the highest scores. 
We score reviews with a binary classifier that inputs review semantic representations. The representations are computed in two steps. First, we independently encode reviews word by word, then compute the weighted average of word representations. Second, we pass all $r_{1:N}$ averaged representations through another encoder (contextualizer) to capture review interdependence features. Details can be found in Appendix~\ref{app:prior_architecture}.



\section{Experimental Setup}
\label{sec:experimental_setup}

\subsection{Data Preprocessing}
\label{sec:data_preprocessing}

In our experiments, we used a preprocessed version of the dataset described in Sec.~\ref{sec:dataset}. First, we set the full review set size $N$ to 100 maximum reviews, and the review subset size $K$ was set to 10 entries. Further, we split the dataset to 26660, 3302, and 3362 summaries for training, validation, and testing, respectively. For our models training, verdicts, pros, and cons were joined to one sequence with a separator symbol indicating boundaries.

\subsection{Baselines}
\label{sec:baselines}

\textsc{LexRank} \citep{erkan2004lexrank} is an unsupervised extractive graph-based model that selects sentences based on graph centrality. Sentences represent nodes in a graph whose edges are weighted with tf-idf. 

\textsc{MeanSum} \citep{chu2019meansum} is an unsupervised abstractive summarization model which treats a summary as a structured latent state of an auto-encoder trained to reconstruct reviews of a product. 

\textsc{Copycat} \citep{brazinskas2020-unsupervised} is the state-of-the-art unsupervised abstractive summarizer with hierarchical continuous latent representations to model products and individual reviews.

\textsc{Random}: here we split all $N$ reviews by sentences, and randomly selected 3, 7, 4 sentences for verdicts, pros, and cons, respectively.

\textsc{ExtSum}:
we created an extractive summarizer trained on our data. First, we used the same ROUGE greedy heuristic as in \citet{liu2019text} to sequentially select summarizing verdict, pro, and con sentences from the full set of reviews using the actual gold summary (\textsc{Oracle}). Further, we trained a model, with the same architecture as the prior in Sec.~\ref{sec:fitting_prior}, to predict sentence classes. More details can be found in Appendix~\ref{app:extractive_summarizer}.

\subsection{Alternative Review Selectors}
\label{sec:alt_review_selectors}

To better understand the role of review selection, we trained the same encoder-decoder summarizer as in \ours{} but with two alternative selectors.

\paragraph{Random reviews}
\label{sec:random_reviews}
We trained and tested on random review subsets (\textsc{RandSel}). Here, review subsets were re-sampled at each training epoch. 


\paragraph{ROUGE-1 top-k}
\label{sec:rouge_topk}
We produced review subsets based on review-summary ROUGE-1 R scores (\textsc{R1 top-k}) for training.\footnote{We tried but were not able to obtain better results by turning the scores into a distribution and sampling from it, so we used the deterministic strategy in the main experiments.} Specifically, we computed the scores for each pair, and then selected $K$ reviews with highest scores to form the subset. To select reviews in test time, we trained a selector as in Sec.~\ref{sec:fitting_prior}.



\subsection{Experimental details}
\label{sec:exp_details}

Below we briefly describe model details; more information can be found in Appendix~\ref{app:experimental_details}.

\paragraph{Summarizer}

We used the Transformer encoder-decoder architecture \citep{vaswani2017attention} initialized with base BART \citep{lewis2019bart}, 140M parameters in total. Reviews were independently encoded and concatenated states of product reviews were attended by the decoder to predict the summary as in \citet{brazinskas2020few}. We used ROUGE-L as the stopping criterion. Summary generation was performed via the beam search of size 5 and with 3-gram blocking~\citep{paulus2017deep}.

\paragraph{Posterior}
For the inference network in Sec.~\ref{sec:approximate_posterior}, we used a simple non-linear two-layer feed-forward network with 250 hidden dimensions. The model consisted of 95k parameters. The network inputs 23 pre-computed features. For instance, ROUGE-1 and -2 scores between each review and the summary, and each review and other reviews in the full set. Similar to \citet{ni2019justifying}, we tagged fine-grained aspect words to compute precision and recall scores between reviews and the summary, and used them as features. 

\paragraph{Prior}
For the parametrized prior in Sec.~\ref{sec:fitting_prior}, we used fine-tuned encoders on the end-task from both \ROUGE{} and \ours{}. For the contextualizer we used a cold-start Transformer encoder with 2 layers and 8-head attention mechanisms. For score networks, we used 2 hidden layer feed-forward networks with the ReLU non-linearities and 100 hidden dimensions. Dropouts at each layer were set to 0.10. In total, the model had 97M parameters. The details of the architecture can be found in Appendix~\ref{app:prior_architecture}.

\paragraph{Pros and cons classification}
\textsc{Copycat} and \textsc{MeanSum} are not specifically designed for \pc{} generation. Therefore, we used a separately trained classifier to split each summary to pros and cons.

\paragraph{Extractive summarizer}
We used a pre-trained BART encoder, and 100 hidden states for 1 layer score feed-forward network with ReLU, and 0.1 dropout. The   contextualizer had one layer, and the final score feed-forward had 100 hidden dimensions, 0.1 dropout, with layer normalization before logits are computed. We trained the model for 5 epochs, with 1e-05 learning rate.
 
\paragraph{Automatic evaluation}
We separately evaluated verdicts, pros, and cons with the standard ROUGE package~\citep{lin2004rouge}\footnote{We used a wrapper over the package \url{https://github.com/pltrdy/files2rouge}.}, and report F1 scores.

\paragraph{Human evaluation}
To assess content support, we randomly sampled 50 products, generated summaries, and hired 3 workers on Amazon Mechanical Turk (AMT) for each HIT. To ensure high quality submissions, we used qualification tasks and filters. More details can be found in Appendix~\ref{app:human_evaluation_setup}.

\paragraph{Hardware}
All experiments were conducted on 4 x GeForce RTX 2080 Ti.

\section{Results}
\label{sec:results}


\subsection{Automatic Evaluation}
\label{sec:automatic_eval}

\begin{table*}[h!]
    \centering
    \begin{tabular}{l | c  c  c | c  c  c | c  c  c }
    \multicolumn{1}{c}{} & \multicolumn{3}{c}{\textbf{Verdict}} & \multicolumn{3}{c}{\textbf{Pros}} & \multicolumn{3}{c}{\textbf{Cons}}\\\thickhline
    & R1 & R2 & RL & R1 & R2 & RL & R1 & R2 & RL \\
    \thickhline
    \textsc{Oracle} & 38.14 & 11.76 & 31.50 & 37.22 & 10.53 & 33.50 & 34.09 & 10.75 & 29.66 \\
    \textsc{Random} & 13.12 & 0.82 & 10.85 & 14.29 & 1.04 & 13.02 & 9.91 & 0.72 & 8.77 \\
    \thickhline
    \textsc{LexRank} & 15.12 & 1.84 & 12.60 & 14.12 & 1.50 & 12.81 & 8.28 & 0.82 & 7.24 \\
    \textsc{MeanSum} & 13.78 & 0.93 & 11.70 & 10.44 & 0.63 & 9.55 & 5.95 & 0.45 & 5.29 \\
    \textsc{Copycat} & 17.05 & 1.78 & 14.50 & 15.12 & 1.48 & 13.85 & 6.81 & 0.82 & 5.89 \\
    \textsc{ExtSum} & 18.74 & 3.01 & 15.74 & 19.06 & 2.47 & 17.49 & 11.63 & 1.19 & 10.44 \\
    \thickhline
    \textsc{RandSel} & 23.25 & 4.75 & 17.82 & 20.26 & 3.60 & 18.52 & 13.59 & 2.32 & 11.86 \\
    \textsc{RandSel*} & 23.95 & 5.16 & 18.49 & 21.06 & 3.94 & 19.31 & 13.78 & 2.35 & 12.10 \\
    \textsc{R1 top-k} & 23.43 & 4.94 & 18.52 & \textbf{22.01} & 3.94 & \textbf{19.84} & 14.93 & 2.57 & 12.96 \\
    
    \ours{} & \textbf{24.33} & \textbf{5.29} & \textbf{18.84} & 21.29 & \textbf{4.00} & 19.39 & \textbf{14.96} & \textbf{2.60} & \textbf{13.07} \\
    
    \thickhline
    \end{tabular}
    \caption{Test set ROUGE F1 scores on verdict, pros and cons. The last block shows review selection variants, where \textsc{RandSel}* was trained on random review subsets but tested on \ours-selected subsets.}
    \label{table:auto_res}
\end{table*}


The results in Table~\ref{table:auto_res} suggest that the supervised models substantially outperform the unsupervised ones. Also, all supervised abstractive summarizers outperform \textsc{ExtSum}, suggesting recombining information from reviews into fluent text is beneficial. 
Among the summarizers with the review selectors, \ours{} yields the best results on verdicts and cons. Although, we noticed that \ours{} generates shorter pros than \ROUGE{}, which may harm its scores~\citep{fan-etal-2018-controllable}\footnote{\ROUGE{} and \ours{} generate 31.95 and 27.14 words on average, respectively. }. Further, when random reviews were used both in training and testing (\textsc{RandSel}), the results are substantially lower. On the other hand, when review subsets were produced by \ours{} and summarized by \textsc{RandSel} (marked with `*'), we observed a substantial increase in all the scores. This suggests the importance of deliberate review selection in test time. In general, all models yield higher scores on pros than cons, which is expected as most reviews
are positive  (on average 4.32/5) and it is harder for the model to find negative points in input reviews.




\subsection{Content Support}
\label{sec:content_support}


    
Generating input faithful summaries is crucial for practical applications, however, it remains an open problem in summarization~\citep{maynez2020faithfulness, fabbri2020summeval, wang2020asking}. Moreover, ROUGE scores were shown not always be reliable for the content support assessment~\citep{tay2019red, brazinskas2020-unsupervised}. Therefore, we evaluated generated summary sentences via AMT, as in \citet{brazinskas2020-unsupervised}, using the following options. 

\textit{Full
  support}: all the content is reflected in the reviews;
\textit{Partial support}: only some content is reflected in the
reviews; \textit{No support}: content is not reflected in the reviews.



\begin{table*}
\centering
\begin{tabular}{l | c  c  c | c  c  c | c  c  c}
    \multicolumn{1}{c}{} & \multicolumn{3}{c}{\textbf{Verdict}} & \multicolumn{3}{c}{\textbf{Pros}} & \multicolumn{3}{c}{\textbf{Cons}}\\\thickhline
    & Full$\uparrow$ & Partial$\downarrow$  & No$\downarrow$ & Full$\uparrow$ & Partial$\downarrow$ & No$\downarrow$ & Full$\uparrow$ & Partial$\downarrow$ & No$\downarrow$ \\ \thickhline
    \textsc{RandSel} & 28.96 & 45.90 & 25.14 & 38.62 & 29.10 & 32.28 & 14.92 & 14.60 & 70.48 \\
    \textsc{RandSel}* & 50.79 & 31.75 & 17.46 & 50.62 & 22.96 & 26.42 & 16.84 & \textbf{13.75} & 69.42 \\
    \ROUGE{} & 55.21 & 31.77 & 13.02 & 56.07 & 26.61 & 17.31 & 33.33 & 27.78 & 38.89 \\
    \ours{} & \textbf{66.08} & \textbf{25.15} & \textbf{8.77} & \textbf{70.21} & \textbf{17.99} & \textbf{11.80} & \textbf{38.41} & 29.21 & \textbf{32.38} \\ \thickhline
\end{tabular}
\caption{Human evaluated content support. Percentages are based on summary sentences. \textsc{RandSel}* was trained on random review subsets but tested on \ours{} selected subsets.}
\label{table:content_support}
\end{table*}

First, we observed that random reviews in training and testing (\textsc{RandSel}) lead to summaries with a significant amount of hallucinations. Further, when \textsc{RandSel} summarizes reviews chosen by \ours{}'s selector (`the prior')  -- indicated by `*' -- the content support is still substantially lower than with \ours{}. This demonstrates that having a selection component is necessary not only at test time but also in training; without it, the model does not learn to be faithful to input reviews. Lastly, \ours{} generates substantially more input faithful summaries than \ROUGE{}.



\subsection{Posterior-Selected Review Subsets}
\label{sec:posterior_selected_review_subsets}


We performed extra experiments to understand why \ours{} model performs better than \ROUGE{}. Recall, their difference is only in the review selector used in training. \ours{} learns a neural model as the posterior, whereas \ROUGE{} relies on a ROUGE-1 heuristic. We hypothesize that \ours{} exploits more expressive features (beyond ROUGE-1) to select reviews that are more relevant to the summary, helping \ours{} to learn a stronger model, less prone to hallucinations.

In order to validate this, in Table~\ref{table:posterior_res} we show their results on the test set but in the training regime, i.e. with reviews selected while accessing the actual gold summary. As in training, \ROUGE{} uses the ROUGE-1 heuristic, while \ours{} relies on the learned posterior. Naturally, both methods obtain stronger scores in this artificial set-up (Table~\ref{table:posterior_res} vs. Table~\ref{table:auto_res}). What is more interesting is that \ours{} is considerably stronger than \ROUGE{}, suggesting that the \ours's selection component indeed chooses more relevant reviews.

\begin{table}[h!]
    \centering
    \begin{tabular}{l | c c c}
    \multicolumn{1}{c}{} & \textbf{Verdict} & \textbf{Pros} & \textbf{Cons} \\\thickhline
    & RL & RL & RL \\
    \thickhline
    \textsc{R1 top-k} & 19.38 & \textbf{21.09} & 13.26 \\
    \ours{} & \textbf{20.44} & 20.79 & \textbf{14.40} \\ 

    \thickhline
    \end{tabular}
    \caption{Test set ROUGE F1 scores when review selection is guided by the gold summary.}
    \label{table:posterior_res}
    \vspace{-2mm}
\end{table}

Lastly, to rank each feature `importance', we estimated the mutual information (MI)~\citep{kraskov2004estimating, ross2014mutual} between the posterior input features and the binary decision to select a review, as in Sec.~\ref{sec:fitting_prior}. We found that besides review-vs-summary ROUGE-1 and -2 scores, the posterior uses fine-grained aspect features, and review-vs-all-reviews ROUGE scores (quantifying the uniqueness of each review). See also Appendix~\ref{app:posterior_features}.

\section{Related Work}
\label{sec:related_work}

Due to a lack of annotated data, extractive weakly-supervised opinion summarization has been the dominant paradigm. 
\textsc{LexRank}~\citep{erkan2004lexrank} is an unsupervised extractive model.
\textsc{Opinosis}~\citep{ganesan2010opinosis} does not use any supervision and relies on POS tags and redundancies to generate short opinions. Although, it can recombine fragments of
input text, it cannot generate novel words and phrases and thus produce coherent abstractive summaries. Other earlier approaches \citep{gerani2014abstractive, di2014hybrid} relied on text planners and templates, which restrict the output text. A more recent method of \citet{angelidis2018summarizing} applies multiple specialized models to produce extractive summaries. 
More recently, there has been a spark of interest in unsupervised abstractive opinion summarization. Such models include \textsc{MeanSum}~\citep{chu2019meansum}, \textsc{Copycat}~\citep{brazinskas2020-unsupervised}, \textsc{DenoiseSum}~\citep{amplayo2020unsupervised}, \textsc{OpinionDigest}~\citep{suhara-etal-2020-opiniondigest}, and  \textsc{CondaSum}~\citep{amplayo-lapata-2021-informative}.


Our work is related to the extractive-abstractive summarization model~\citep{chen2018fast} that selects salient sentences from an input document using reinforcement learning. They assume one-to-one mapping between extracted and summary sentences for news. In opinion summarization, however, we often need to fuse  user opinions expressed in multiple reviews. Lastly, unlike their model, our selector and summarizer are trained jointly to predict the summary using a differentiable loss. Also, our model is related to the unsupervised paraphrasing \textsc{MARGE} model \citep{lewis2020pre}, where the decoder has a modified attention mechanism accounting for the target-source document similarity. However, in their approach, the actual selection of relevant documents is performed offline via heuristics. This, in turn, makes it non-differentiable and over-reliant on the modified attention mechanism. We, however, learn the selector (posterior) jointly with summarizer, and select reviews in the online regime.

An alternative to review subsets selection are more memory and computationally efficient attention mechanisms~\citep{Beltagy2020Longformer, pasunuru-etal-2021-efficiently}. However, it is unclear what relationship exists between attention weights and model outputs~\citep{jain-wallace-2019-attention}, thus, making it harder to offer evidence for generated summaries. In our case, the summarizer relies only on a selected subset and generates summaries faithful to its content. 

In general, in news summarization, which is a more mature branch, large datasets are commonly obtained from online resources~\citep{sandhaus2008new, hermann2015teaching, grusky2018newsroom, narayan2018don, fabbri2019multi}. The most relevant dataset is \textsc{MultiNews} \citet{fabbri2019multi}, where journalist-written summaries are linked to multiple news articles. The most similar opinion summarization dataset \textsc{Space}~\citep{angelidis2020extractive} contains 1050 summaries produced for 50 hotels by crowd-sourcing.

\section{Conclusions}
\label{sec:conclusions}

In this work, we introduce the largest multi-document abstractive dataset for opinion summarization. The dataset consists of verdicts, pros and cons, written by professional writers for more than 31,000 Amazon products. Each product is linked to more than 320 customer reviews, on average. As standard encoding-decoding is computationally challenging, we perform summarization with an integrated component that selects smaller review subsets. We conclude that the `naive' selection of random reviews leads to content infidelity (aka hallucinations) and present \ours{} that learns to select and summarize reviews end-to-end. The model is computationally efficient, scaling to large collections. 
Its summaries result in better ROUGE scores and are better supported by input reviews.


\subsection{Ethics Statement}
\label{sec:ethics}

\paragraph{Human Evaluation}
We used a publicly available service (Amazon Mechanical Turk) to hire voluntary participants, requesting native speakers of English. The participants were compensated above minimum hourly wage both in the USA and the UK, the self-reported locations of the participants. 

\paragraph{Dataset}

The dataset was collected and used in accordance to the non-commercial purpose. The dataset is intended for non-commercial and educational purposes only. It will be made available free of charge for these purposes without claiming any rights, similar to \citep{grusky2018newsroom}. To maintain privacy, all summary writers are anonymized.  


\section*{Acknowledgments}

We would like to thank members of Edinburgh NLP group for discussion. Also, we would like to thank Serhii Havrylov, Wilker Aziz (UvA, ILLC), and anonymous reviewers for their input and helpful comments. We gratefully acknowledge the support of the European Research
Council (Titov: ERC StG BroadSem 678254; Lapata: ERC CoG TransModal 681760) and the Dutch National Science Foundation (NWO VIDI 639.022.518).

\bibliographystyle{acl_natbib}
\bibliography{anthology,custom}

\newpage
\section{Appendices}
\label{sec:appendix}

\subsection{REINFORCE vs Gumbel-Softmax}
\label{app:RI_vs_gs}

In our experiments, we used REINFORCE \citep{williams1992simple} instead of a straight-trough Gumbel-Softmax estimator~\citep{jang2016categorical}, which is a popular alternative. In our case, we need to sample without replacement each $\hat r_k$ from the collection $r_{1:N}$. The Gumbel-Softmax, requires to relax the system for `soft' samples used in the backward pass. In this way, the system is updated considering all possible assignments to the categorical variable, even though only one was sampled in the forward pass. For instance, one could encode all reviews $r_{1:N}$ and weigh their word contextualized representations to obtain each $\hat r_k$. However, this is a computationally expensive and memory demanding operation. On the other hand, REINFORCE does not require this relaxation, and the encoder is exposed only to one possible assignment to each $\hat r_k$, both in the forward and backward pass.


\subsection{Extractive Summarizer}
\label{app:extractive_summarizer}

As mentioned in Sec.~\ref{sec:baselines}, our extractive summarizer had the same architecture as the prior in Sec.~\ref{sec:fitting_prior}. We independently encoded sentences from reviews, contextualized them, and computed their distributions for 4 classes. 

In training, we considered up to 550 sentences, where only up to 16 have positive labels (4, 8, 4 for verdicts, pros, cons, respectively) marked by \textsc{Oracle}. However, this resulted in label imbalance, where, in training, the model is incentives to ignore positive labels~\citep{li2019dice}. However, in test time, we care about positive instances only. To counter this problem, we scaled each positive class loss by 50. In this way, the model is forced to prioritize the positive classes more. 

At test time, we sequentially selected top-k summarizing sentences for verdicts, pros, and cons. To make each sentence selected either for verdict, pros, or cons, we were sequentially excluding selected sentences from the pool of candidates.


\subsection{Prior Score Function}
\label{app:prior_architecture}

\begin{figure}[t!]
    \centering
    \includegraphics[width=0.45\textwidth]{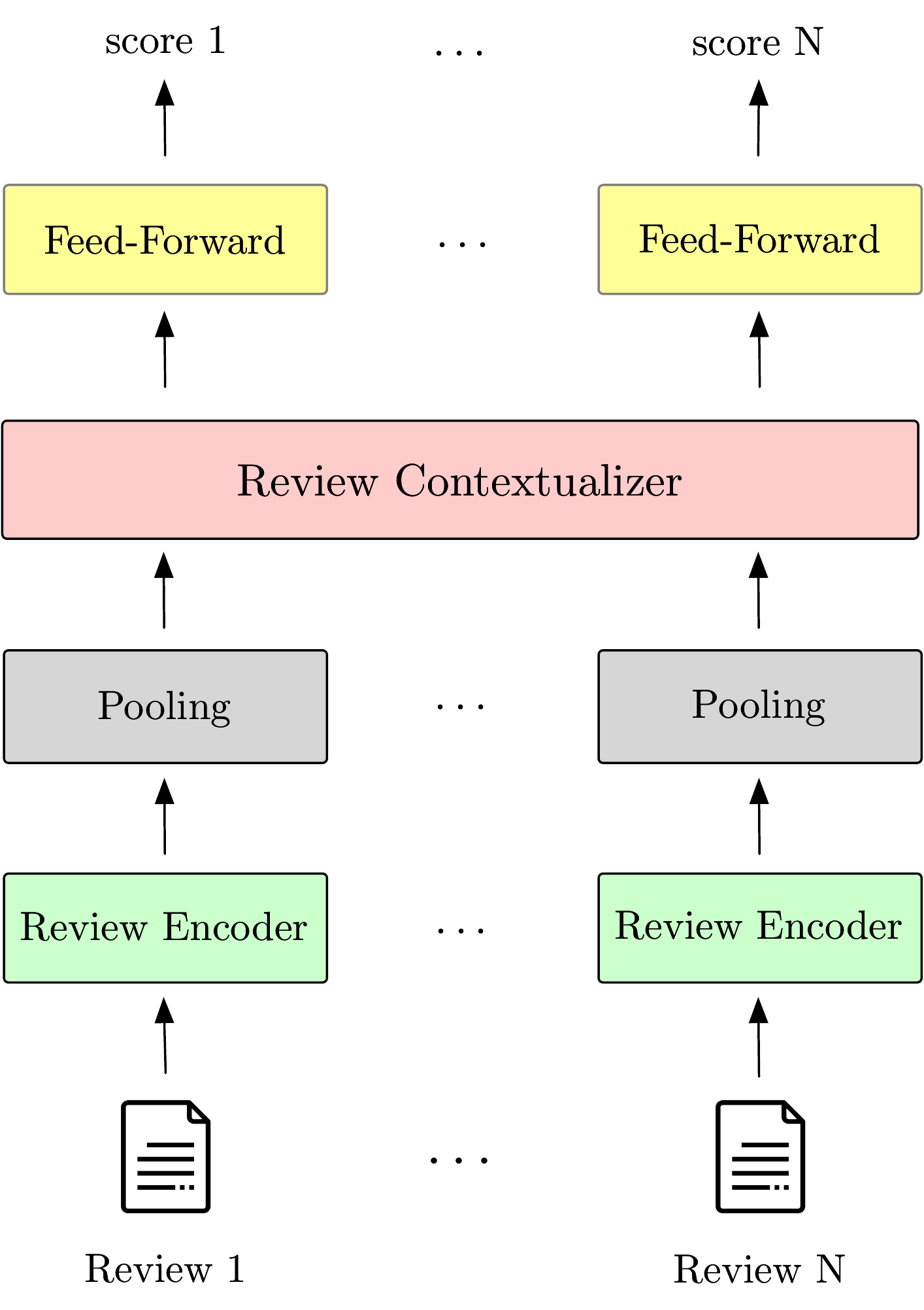}
    \caption{Architecture of the prior score function.}
    \label{fig:prior_score_function}
\end{figure}

Below we describe the architecture of the score function used in Sec.~\ref{sec:fitting_prior}, and it is also illustrated in Fig.~\ref{fig:prior_score_function}. First, we initialized with a fine-tuned review encoder from the summarizer that was trained using a review selector (i.e. \ours{} or \ROUGE{}). The encoder produces contextualized word representations for each review independently. The word representations are obtained from the last Transformer layer. Then, we computed the weighted average of these representations to get the review representation. Further, we passed the review representations through another encoder that contextualizes them by attending representations of other reviews in the collection. Finally, we projected the outputs to scores.

\subsection{Human Evaluation Setup}
\label{app:human_evaluation_setup}

To perform the human evaluation experiments described in Sec.\ref{sec:content_support}, we hired workers with 98\% approval  rate, 1000+ HITS, from the USA and UK, and the maximum score on a qualification test that we had designed. We payed them 17.25 \$ per hour, on average. The task was a minimal version of the actual HIT, where we could test that workers correctly understood the instructions. Also, we asked them if they were native English speakers.

\subsection{Experimental Details}
\label{app:experimental_details}

\paragraph{Summarizer}

We used the Transformer encoder-decoder architecture \citep{vaswani2017attention} initialized with base BART \citep{lewis2019bart}, 140M parameters in total. Reviews were independently encoded and concatenated states of product reviews were attended by the decoder to predict the summary as in \citet{brazinskas2020few}. We used trainable length embeddings, and BPE \citep{sennrich2015neural} vocabulary of 51,200 subwords. Subword embeddings were shared across the encoder and decoder for regularization \citep{press2016using}. For summary generation, we used beam search with the size of 5 and 3-gram blocking \citep{paulus2017deep}. Parameter optimization was performed using Adam \citep{kingma2014adam} with 5,000 warm-up steps. We trained \ours{}, \ROUGE{}, and \textsc{RandSel} for 8, 8, and 9 epochs, respectively. All with the learning rate of 3e-05.

\paragraph{Posterior}
For the inference network in Sec.~\ref{sec:approximate_posterior}, we used a simple two-layer feed-forward, 250 hidden dimensions, with the tanh non-linearity and layer normalization before a linear transformation to scores. The model consisted of 95k parameters. We used 23 static features by treating verdicts and \pc{} as separate summaries. For instance, ROUGE-1 and -2 scores between each review and the summary, and each review and other reviews in the full set. Similar to \citet{ni2019justifying}, we tagged fine-grained aspect words to compute precision and recall scores between reviews and the summary, and used them as features. Full details about features can be found in Appendix~\ref{app:posterior_features}.
Lastly, we used 3 samples for the expectation estimation in Eq.~\ref{eq:rec_grad} and 3 samples to compute the baseline.

\paragraph{Prior}
For the parametrized prior in Sec.~\ref{sec:fitting_prior}, we used fine-tuned encoders on the end-task from both \ROUGE{} and \ours{}. For the contextualizer we used a cold-start Transformer encoder with 2 layers, and 8-head attention mechanisms. For score networks, we used 2 hidden layer feed-forward networks with the ReLU non-linearities, and dropouts set to 0.10. In total, the model had 97M parameters. We trained the one for \ours{} and \ROUGE{} for 5 and 4 epochs, respectively, with the 1e-05 learning rate and 5,000 warmup steps. The details of the architecture can be found in Appendix~\ref{app:prior_architecture}.

\subsection{Aspect-based Metric}
\label{app:aspect_based_metric}

In addition to standard unweighted word-overlap metrics commonly used to analyze datasets~\citep{grusky2018newsroom, fabbri2019multi}, we also leveraged an aspect specific metric. Similar to \citet{ni2019justifying}, we applied a parser~\citep{zhang2014explicit} to the training set to yield \textit{(aspect, opinion, polarity)} tuples. From the tuples, we created a lexicon, which contains fine-grained aspect keywords, such as battery life, screen, resolution, etc. In addition, to reduce noise, we manually cleaned the lexicon from aspect unrelated keywords, resulting in 2,810 entries.
Further, we used the lexicon to automatically tag aspect keywords in text. Lastly, we computed \textit{aspect precision} (AP) and \textit{aspect recall} (AR) scores by comparing two sequences. These scores were used as features in \ours{}.

\subsection{Posterior Features}
\label{app:posterior_features}



\begin{table}[t!]
    \centering
    \begin{tabular}{l | c}
    Feature & MI\\
    \thickhline
    R2-R($\hat r$, $pc$) & 0.0634 \\
    R1-R($\hat r$, $pc$) & 0.0564 \\
    R2-P($\hat r$, $pc$) & 0.0523 \\
    R1-R($\hat r$, $v$) & 0.0489 \\
    R2-R($\hat r$, $v$) & 0.0449 \\
    R2-P($\hat r_k$, $r_{-k}$) & 0.0411 \\
    R2-P($\hat r$, $v$) & 0.0405 \\
    AR($\hat r$, $pc$) & 0.0353 \\
    R1-R($\hat r_k$, $r_{-k}$) & 0.0346 \\
    AP($\hat r$, $pc$) & 0.0331 \\
    R2-R($\hat r_k$, $r_{-k}$) & 0.0313 \\
    R1-P($\hat r$, $pc$) & 0.0266 \\
    R1-P($\hat r$, $v$) & 0.0208 \\
    AP($\hat r_k$, $r_{-k}$) & 0.0190 \\
    AR($\hat r_k$, $r_{-k}$) & 0.0173 \\
    LD($\hat r$, $pc$) & 0.0167 \\
    AP($\hat r$, $v$) & 0.0151 \\
    LD($\hat r$, $v$) & 0.0146 \\
    R1-P($\hat r_k$, $r_{-k}$) & 0.0138 \\
    AR($\hat r$, $v$) & 0.0135 \\
    AD($\hat r$) & 0.0106\\
    AD($v$) & 0.0005 \\
    AD($pc$) & 0.0003 \\ \thickhline
    \end{tabular}
    \caption{Full list of features sorted by their mutual information to a review being selected to the subset binary variable.}
    \label{table:posterior_features_mi_long}
    \NoGap{}
\end{table}

In total, we used 23 continuous features computed for each tuple ($s$, $\hat r_k$, $\hat r_{1:N}$). We can group features into three categories. The first ones were computed for a sequence standalone. The second ones were computed as $f(\hat r_k, s)$ where $\hat r_k$ is the current review (hypothesis) and $s$ is the summary (reference). The last ones were computed with respect to other reviews as $f(\hat r_k, r_{-k})$, where $r_{-k}$ (reference) are all reviews except $\hat r_k$ (hypothesis). We also treated verdicts ($v$) and \pc{} ($pc$) as separate sequences. 

Aspect precision, recall, and density are calculated by leveraging the lexicon presented in Appendix~\ref{app:aspect_based_metric}. Additionally, aspect density (AD) was computed as the number of unigram aspect keywords, divided by the number of unigrams in a sequence. Finally, length difference LD($\cdot$, $\cdot$) was computed as the difference of two normalized lengths. The Normalization was performed by the maximum sequence length division.

To gain a deeper insight into the \ours{} posterior's inner-workings, we analyzed feature importance for including a review in the subset. Same as in Sec.~\ref{sec:fitting_prior}, we used the trained posterior to create a binary tagging dataset. Further, we estimated the mutual information (MI)~\citep{kraskov2004estimating, ross2014mutual} between the posterior input features and the binary decision variable. This allowed us to identify the dependency strength between each feature and the variable.

As features are computed separately for verdicts and \pc{}, we observed that features for \pc{} ($pc$) have higher MI than for verdicts ($v$), which suggests that review selection is guided by \pc{} more than by verdicts. Second, fine-grained aspect keyword based scores (AP and AR) also have high MI for $pc$. This is unsurprising, as \pc{} are often more detailed, making them less predictable based on the prefix, thus the model favours reviews with matching aspect keywords. Lastly, the ROUGE scores computed against other reviews in the collection $r_{-k}$ have high MI. This indicates reliance on global statistics computed based on the whole set of reviews.

\subsection{Error Analysis}
\label{app:error_analysis}

Human written \pc{}, besides summarizing customer opinion expressed in reviews, can contain details that can often be found in the product meta data. For example, users rarely mention that the same product comes in different colors. Consequently, the decoder is trained to predict such phrases based on the prefix instead of the input reviews, as shown in Example~\ref{table:example_1}. We further observed that cons are harder, in general, to align to reviews. 

Human-written summaries sometimes contain customer opinion quantification expressed in phrases, such as "some users" and "a few customers". We observed this to be challenging for the summarizer to generate accurately as shown in Example~\ref{table:example_2}. Especially, it applies to cons that summarize opinions of a small number of users. Logically, such reviews are hard to retrieve from a large collection in training, consequently, the model learns to rely on local statistics (the prefix).  
Overall, quantification of user opinions adds an additional layer of complexity for the decoder as besides generating summaries that are content supported in terms of opinions, it needs to quantify the them correctly. This is an interesting future direction for abstractive opinion summarization. 
Lastly, we observed that online users, in their reviews, sometimes compare the product to other products on the market. This, in turn, can confuse the model and 
make it generate the summary that contains fragments describing another product. Occasionally, we observed such mistakes in the output.



\begin{table*}
    \centering
 	\footnotesize 
    \begin{tabular}{  >{\centering\arraybackslash} m{1.5cm} m{12cm}}
 \thickhline
    Verdict & \tablegap \textcolor{cyan}{A comprehensive study guide} for those who are new to the ASWB exam. \tablegap \\ \hline
    Pros & \tablegap \textcolor{myGreen}{Offers a variety of practice questions to help you get the most out of the exam}.
        \textcolor{red}{Offers an easy-to-understand overview of the test} and how it works. \tablegap \\ \hline
    Cons & \tablegap \textcolor{blue}{The practice questions are not as detailed as the actual exam, and some questions may not be relevant to the actual questions}.
 \tablegap \\ 
     \thickhline
    Review 1 & \tablegap This review guide claims to reflect the 2018 blueprint of the ASWB exam.
        However, all ethics questions refer to a 2006 version of the NASW Code of Ethics.
        The code of ethics is a substantial part of the exam and many of the questions and answer explanations do not reflect what will be on the test.
        It is not worth the money.
        \tablegap \\ \hline
    Review 2 & \tablegap I passed my LMSW test on the first try with this as the only study material!!!
        I would definitely recommend the book for its content and practice test.
        \textcolor{blue}{I will say that the actual test is very different from this practice test} in the book as the real tests involves more questions what do you do FIRST and what do you do NEXT? This guide makes it pretty easy to narrow down to two answers whereas the actual test is not that easy.
        Study all the content and know it!
        Supervisory and ethics played a big part in this test.
        Be more prepared for simple application questions than objective content...
        \tablegap \\ \hline
    Review 3 & \tablegap I used this book as my primary study material for the LMSW licensing exam.
        While it is a \textcolor{cyan}{thick book with a lot of information, it was helpful in preparing for the exam}.
        There were some questions on the exam that were not in the book, but I still passed the exam with the information I studied from the book.
        While the exams are not all the same, I can not guarantee the same results for everyone who uses this book, but I do not have any negative reviews.
        I bought the book used so I did not utilize the app that comes with it.
        \tablegap \\ \hline
    Review 4 & \tablegap I just graduated with my MSW in May, and studied with this book as well as the pocket prep app for about a month.
        This book is a \textcolor{red}{great comprehensive overview of material} that we have all learned, and was great for reviewing.
        \textcolor{myGreen}{The practice questions were also helpful in figuring out HOW the exam wants you to answer}.
        I passed the LMSW exam on my first try today!
        I will definitely buy the clinical version when I take that in a few years!
        \tablegap \\ \hline
    Review 5 & \tablegap 
            I read through this book and utilized the practice exam at the end.
        While this book does go over some foundational content which is applicable to the exam, \textcolor{blue}{overall the content is redundant and irrelevant}.
        The practice exam in the back of \textcolor{blue}{this book is extremely different than the practice exam offered through the ASWB or the actual exam itself}.
        Multiple licensed social workers I have spoken to have stated that the practice exam offered through the ASWB was the most helpful thing in preparing for the actual exam and understanding how its questions are formatted, which I 100\% agree with.
        For the money I spent on this book, it was disappointingly ineffective as an exam prep tool.
    \tablegap \\ \hline
    Review 6 & \tablegap Yes, I did it.
        I have an LCSW(not to be confused with the LCSW-clinical license as in MA the 'C' stands for certified).
        \textcolor{cyan}{With this comprehensive and thurough study guide} \textcolor{myGreen}{i was able to pass my exam the first time}.
        I felt so prepared after using this.
        I would say, pair this study guide with an online prep app, as the exam is computer based and using a phone app you will get into the patterning needed to succeed in the exam.
    \tablegap \\ \hline
    Review 7 &  \tablegap This study guide for the ASWB exam is great.
        It has a lot of review material and a practice test.
        There is also a code to put on a phone/tablet.
        I used this and passed on the first time.
        I also used the BSW guide for that exam and passed on the first time.
        A must have for anyone looking to pass the Master's Exam.
        \tablegap \\ \hline
    Review 8 & \tablegap This was a super purchase!
        It offers excellent tips and strategies to prepping for this challenging exam.
        \textcolor{red}{It conditioned me to understand the method of the questions and not just knowledge.}
        I just passed the first time!
        This was incredible because I trained in the UK and not USA.
        This study kit prepared me to pass what better review can one give?
        \tablegap \\ \hline
    Review 9 & \tablegap I passed!!
        \textcolor{cyan}{This was a great study guide for me.
        I was intending to read the whole thing but it was a lot so I went through the table of contents and highlighted the sections I wanted to study}.
        It also helped to read it and write it down for memorization.
        The practice test was hard but it really tests you on your knowledge so don't take it until you are ready.
        I used this and a few other practice exams.
        \tablegap \\ \hline
    Review 10 & \tablegap  
    I cannot attest to results as of yet.
        But I can say that \textcolor{red}{the book has a very organized layout and presents information about how the test is setup which provides great insight for one's approach to testing.}
        \tablegap
    \\ \thickhline
    \end{tabular}
    \caption{Example summary generated by \ours{} with color highlighted content alignment.}
    \label{table:example_4}
\end{table*}

\begin{table*}
    \centering
 	\footnotesize 
    \begin{tabular}{  >{\centering\arraybackslash} m{1.5cm} m{12cm}}
 \thickhline
    Verdict & \tablegap If you're looking for a set of glass containers that are both BPA-free and dishwasher safe, this is the one to get. \tablegap \\ \hline 
    Pros & \tablegap Glass containers come in a variety of sizes and \textcolor{red}{colors}, so you can find the right size for your needs. \textcolor{cyan}{The lids are easy to open and close}. BPA and \textcolor{red}{phthalate-free}. \tablegap \\ \hline
    Cons & \tablegap The containers are on \textcolor{orange}{the smaller side, which can make them difficult to store in the microwave}. \tablegap \\ 
     \thickhline
    Review 1 & \tablegap  These containers are fantastic.
        The lips snap on very securely but are extremely easy to remove.
        I do wash the lids by hand b/c they are top rack dishwasher safe only but that's not a big deal for me b/c the glass can go through the dishwasher.
        Plus I am saving dishwashing time anyway b/c I previously stored leftovers in Tupperware that I could not heat in the microwave so would have to transfer to a different dish before heating.
        Now it's a 1 stop shop.
        Variety of sizes are great as well. \tablegap \\ \hline
    Review 2 & \tablegap  These containers are excellent!!
        One of the things I love about them is that I can fill them all the way to the top and it won't spill out when I put the lid on.
        There are a variety of sizes included and I use these every day.
        I only wish they were etched on the bottom with the volume that each bowl can hold.
        But in any case they are worth every penny.
        Glass containers don't discolor in the microwave and you don't have to worry about consuming plastic.
        The lids add a layer of sturdiness to the bowls and I store them in the cupboard with their lids on. \tablegap \\ \hline
    Review 3 & \tablegap  Pro: Glass containers can go into the microwave to reheat leftovers without cooking food oils and colors into plastic.
        Don't use the lids in the microwave.
        And DO follow the instructions and wash before using, definitely!
        Slightly con: The rubber gasket will separate from the lid and stick to the glass container if you snap the lid on while it's slightly wet.
        Maybe if it's completely dry as well!
        But it's easy enough to peel off and reseat in the lid. \tablegap \\ \hline
    Review 4 & \tablegap         Great quality!
        Nice, secure fitting lids.
        It's so easy to know what is being stored in the bowls.
        I have not used them in the freezer.
        I have only used them in the microwave and refrigerator.
        We have used every single bowl at one time or another!
        It's great to have different sizes to accommodate different portions.
        I would like to get a couple of even larger sizes if I can find them!
        This glass won't crack as my plastic containers did (unless I happen to drop!).
        An interlocking system between the bottom of a container and the lid of another, so they would stack more securely, would be REALLY NICE. \tablegap \\ \hline
    Review 5 & \tablegap         These are the perfect size for everything, but I'm sad that the blue rubber isn't staying on the lids at all.
        We are avoiding washing them in dishwasher so they don't get worse.
        Kinda bummed.
        UPDATE 6/3/2018 We received the new and improved 1790 Glass Container Set \& the modifications made by the manufacturer were a home run!
        The lids fit tightly and evenly on the containers that using them is a snap.
        They are so well made.
        Completely air tight and leak proof.
        Very impressed with how quickly the issue was addressed and resolved.
        5 Star product!
        If you're looking for versatile containers that deliver..... look no further.
        They are right here. \tablegap \\ \hline
    Review 6 & \tablegap These are such great dishes!
        I don't eat a lot so they are perfect for single serving cooking and storage.
        Love the way the lids clip in place, making a very good seal to keep the food fresh.
        Baking and cleaning are so easy, just the way I like things, nice and easy.
        Would recommend this set to anyone looking for a set of small versatile baking/storage options! \tablegap \\ \hline
    Review 7 & \tablegap Some of the flaps on \textcolor{cyan}{the lids are a little hard to close}, but I am guessing that has more to do with the fact that they are new more than anything else.
        Overall, this is a good, quality set at a great price.
        Durable in oven and microwave and washes up easily.
        No staining, bubbles or potential melting issues like plastic containers. \tablegap \\ \hline
    Review 8 & \tablegap         I should have read a little closer and counted the actual dishes in the photo - its a NINE piece set unless you count the lids - which you can't store food in.
        They might as well call it a 27 piece set because of the nine lid seals, which is realistically another 'piece'.
        The quality is average along with price when you discover how many actual storage containers are sold.
        So, read the entire description and count the dishes in the photo.. \tablegap \\ \hline
    Review 9 & \tablegap         Great glass based meal containers.
        The caps are plastic, and the air seal is a rubber-like material.
        Works as intended, however the seal part can be separated from the cap and has a tendency to adhere more onto to the glass over the cap.
        If this happens, be careful when separating the seal and container, as it can cause the air seal to rip a bit. \tablegap \\ \hline
    Review 10 & \tablegap   This product is durable and easy to clean.
        I love that it's BPA free and oven, microwave, freezer, and dishwasher safe.
        It's air tight and I haven't had anything leak even when putting liquid in.
        For the price you get, they're great storage containers with versatility in different temperatures.
        I got this for my boyfriend and will probably buy more for myself if I needed glass containers in the future. \tablegap \\ \thickhline
    \end{tabular}
    \caption{Example summary generated by \ours{} with color highlighted errors. As indicated in \textcolor{red}{red}, pros can contain details that one would expect to find in product meta data instead of customer reviews. In \textcolor{orange}{orange}, we indicate a logical mistake. In \textcolor{cyan}{cyan} we indicate a contradiction.}
    \label{table:example_1}
\end{table*}

\begin{table*}
    \centering
 	\footnotesize 
    \begin{tabular}{  >{\centering\arraybackslash} m{1.5cm} m{12cm}}
 \thickhline
    Verdict & \tablegap If you're looking for a reliable, water-resistant, and weather-resistant G-Shock watch, this is the one to get. \tablegap \\ \hline
    Pros & \tablegap Solar powered.
        Water-resistant.
        Includes atomic clock and countdown timer.
        Includes a stopwatch and 5 alarms. \tablegap \\ \hline
    Cons & \tablegap
    \textcolor{red}{Some owners} say the watch is \textcolor{red}{too small} for their needs.
        \textcolor{orange}{A few owners} say it \textcolor{orange}{doesn't have the features of other models}. 
        \tablegap \\
     \thickhline
    Review 1 & \tablegap  It will become your everyday watch and you will enjoy it.
        It does everything in the description and then some.
        It sits nicely on your wrist without looking too big or too small, just right.
        It has stopwatch capabilities along with timer capabilities.
        Its a good digital watch and from what I've seen it can take a beating and still beep on the hour for you.
        Of course there's a couple accessories you can get for it too, including a screen protector and a brace to protect it.
        If you're looking for a good watch that you can drag through the mud and still check the time, look no further!
        \tablegap \\ \hline
    Review 2 & \tablegap  
    This Casio G-Shock adjusted itself to the correct time, day of the week and date as soon as I unboxed it and light struck the solar charger.
        Very easy to set up, unlike some Casio products that require long sequences of button pushing.
        It meets my needs: updates the time and date automatically via an atomic clock signal, is solar powered, is water resistant, and displays the important information at a glance.
        Glad I paid a little more for this model than some of the Casios that have thick operating manuals and lots of button pushing to adjust.
        Well worth the money, and Amazon did a great job with prompt delivery of the correct product on top condition. \tablegap \\ \hline
    Review 3 & \tablegap
    OK - so it won't give you weather, or headings, or your email... but come on, it sets itself to the freakin' atomic clock EVERY SINGLE DAY.
        AND it's shock resistant, and water resistant, and solar powered, and lights itself up when you turn it towards your face, AND it has a stop watch, AND it can tell you the time in multiple time zones. SERIOUSLY, this watch is a sleeper, FANTASTIC value for the money.
        \tablegap \\ \hline
    Review 4 & \tablegap 
    As a purchaser of Casio' G-Shock watches for almost four decades, I have depended on their toughness in rugged environments.
        This latest update has all the whistles and bells of the bigger versions, but more compact.
        The atomic-solar G-Shocks seem to last ten years with accurate time and no problems with the battery.
        I recommend this watch for those who want a minimum of hassle.
        (Note: I rarely use backlight illumination.)
        \tablegap \\ \hline
    Review 5 & \tablegap    
    I like the atomic clock sync and multi-time zone options.
        The UTC time needs to be accurate and easy to get to for celestial navigation.
        This watch has all that plus many other features and, of course, it's nearly impossible to break and self-charging.
        This is the perfect watch for someone that is outdoors or on the ocean for extended periods.
        I highly recommend this for anyone that needs a completely reliable and accurate time piece. 
        \tablegap \\ \hline
    Review 6 & \tablegap
    This is one of the best value for the money G-Shocks out there.
        A homage to the original square G-Shock, this watch combines retro styling with modern technology like automatic time setting.
        Every feature works flawlessly, and it looks and feels good on the wrist.
        If you're strapped for cash go for the 5600 series, but if you can swing it, this is the entry level G-Shock that will leave you wanting for nothing.
        Easy to mod or awesome as is.
        \tablegap \\ \hline
    Review 7 & \tablegap
    It's cheap, reliable, durable, and fashionable.
        It's so light it feels like you're not wearing a wrist watch.
        It tells the time, date, day of the week.
        It adjusts itself to the atomic clock nearest to you.
        It's solar powered.
        You don't have to adjust it or buy a new battery for it (for around a decade!).
        It has a stopwatch and 5 alarms although I don't see myself using them.
        It can quickly adjust to different time zones.
        You can use it it in the dark.
        I don't know if I listed all the advantages...
        \tablegap \\ \hline
    Review 8 & \tablegap         
    I love these watches.
        It is flippin' solar powered with multi-band atomic correction!
        You can't beat that for maintenance-free operation.
        I appreciate the simple design and layout without the cluttered display or bulk of other models. A simple, durable, reliable wrist watch.
        \tablegap \\ \hline
    Review 9 & \tablegap        
    If your interested in a no flash, timeless, simplistic design then this watch is for you.
        This G-Shock is a step up from Casio classic DW-6500.
        It add solar charging, world time, 5 alarms, and the ability to sync with 6 different stations around the world.
        As a fan of Casio's Square G-Shocks this one in particular is one of my favs.
        It keeps that classic look and durability while improving on it.
        \tablegap \\ \hline
    Review 10 & \tablegap  This model is considered a must have by most G-Shock aficionados.
        It's the classic G-Square model, but updated with solar power and atomic clock time sync features.
        It also has the World Time feature.
        It's a rock-solid watch that is nearly indestructible.
        \tablegap \\
    \\ \thickhline
    \end{tabular}
    \caption{Example summary generated by \ours{} with highlighted errors. In this summary, the system incorrectly generated cons with quantifiers.}
    \label{table:example_2}
\end{table*}

\begin{table*}
    \centering
 	\footnotesize 
    \begin{tabular}{  >{\centering\arraybackslash} m{1.5cm} m{12cm}}
    
     \multicolumn{2}{c}{\textbf{Gold}} \vspace{0.5em} \\ \thickhline
    Verdict & \tablegap If you need to keep rodents away from several \textcolor{red}{small places}, this is a good option.
 \tablegap \\ \hline
    Pros & \tablegap
    \textcolor{red}{Available in packs of 1, 2, 3, or 4}. Deters rodents of all kinds.
    Safe to use around kids and \textcolor{red}{pets} - \textcolor{red}{no noticeable sound}. Comes with a built-in night light. Easy to use - just plug in any room where pests are detected. \tablegap \\ \hline
    Cons & \tablegap  
        \textcolor{orange}{Does not repel insects, only rodents}. \textcolor{red}{Takes about a week to work.}
    \vspace{1.0em} \\   
    \multicolumn{2}{c}{\textbf{Generated}} \tablegap \\ \thickhline
    Verdict & \tablegap If you are looking for \textcolor{red}{an inexpensive way} to attract and control rodents, this 3-pack of nightlights is a good choice. \tablegap \\ \hline
    Pros & \tablegap Three nightlights for indoor or outdoor use. \textcolor{red}{Lightweight}, compact, and \textcolor{red}{easy to store}. Can be used indoors or \textcolor{red}{outdoors}. \tablegap \\ \hline
    Cons & \tablegap \textcolor{orange}{Not as effective against rodents} as it is for trapping them. Some reports of the nightlights not working. \vspace{1.0em} \\ 
    \multicolumn{2}{c}{\textbf{Selected Reviews}} \tablegap \\
     \thickhline
    Review 1 & \tablegap  Description and photo showed 2 nightlights. Package arrived with one nightlight and 2 unlit units. Kinda annoying Overall, does what it's designed to do. \tablegap \\ \hline
    Review 2 & \tablegap
    Decent product. Rarely see anything anymore only thing is it messes with the frequency of your belongings like microwave toaster etc my advice is keep them low level but overall nice.
        \tablegap \\ \hline
    Review 3 & \tablegap 
    So far, these have worked great for ants actually. We haven't seen any critters at all.
    \tablegap \\ \hline
    Review 4 & \tablegap
    I can't say how well they work as far as rodents go, but I have yet to see one EVER in my house. I love the little light it illuminates on the floor. It is fantastic at night and gives me piece of mind for little critters.
    \tablegap \\ \hline
    Review 5 & \tablegap 
    We have not had any more mice in our garage!
    \tablegap \\ \hline
    Review 6 & \tablegap 
    These little babies actually seem to work! We live in an area with lots of pests.
    I put 6 of these around the house and we haven't had any issues whatsoever.
    \tablegap \\ \hline
    Review 7 & \tablegap 
    Bought one of these devices and used while at winter home.
    When we returned see no sign of critters where we installed.
    \tablegap \\ \hline
    Review 8 & \tablegap 
    I can't believe it seems to drive them away! Happy Camper.
    \tablegap \\ \hline
    Review 9 & \tablegap 
    The nightlights on all 3 stopped working less than 6 months after I started using them. I called the company and was told that if the nightlights are out then the product isn't working. They advertise that these will last for 3-5 years. Did the product work when the nightlights were working? I don't know because I was still trapping mice in the traps that were nearby!
    \tablegap \\ \hline
    Review 10 & \tablegap 
    Excellent. I put 2 in each room. I haven't seen a trace since.
    \tablegap \\ \thickhline
    \end{tabular}
    \caption{Example summary generated by \textsc{RandSel}, where reviews are randomly sampled in training and testing. Extrinsic hallucinations are marked in \textcolor{red}{red}, intrinsic in \textcolor{orange}{orange}. When random review subsets are used, the decoder is forced to invent `novel' content in training leading to hallucinations in test time.}
    \label{table:randsel_example}
\end{table*}

\end{document}